\documentclass[conference]{IEEEtran}
\IEEEoverridecommandlockouts
\usepackage{cite}
\usepackage{amsmath,amssymb,amsfonts}
\usepackage{algorithmic}
\usepackage{graphicx}
\usepackage{textcomp}
\usepackage{xcolor}
\def\BibTeX{{\rm B\kern-.05em{\sc i\kern-.025em b}\kern-.08em
    T\kern-.1667em\lower.7ex\hbox{E}\kern-.125emX}}
    
\begin{document}

\title{Residual GRU+MHSA: A Lightweight Hybrid Recurrent Attention Model for Cardiovascular Disease Detection}

\author{
\IEEEauthorblockN{
Tejaswani Dash\textsuperscript{1},
Gautam Datla\textsuperscript{2},
Anudeep Vurity\textsuperscript{3},
Tazeem Ahmad\textsuperscript{4},
Mohd Adnan\textsuperscript{5},\\
Saima Rafi\textsuperscript{6}
Saisha Patro\textsuperscript{7}
Saina Patro\textsuperscript{7}
}

\IEEEauthorblockA{\textsuperscript{1}George Mason University, Fairfax, USA; \quad Email- tdash@gmu.edu}
\IEEEauthorblockA{\textsuperscript{2}NJIT, Newark, New Jersey, USA; \quad Email- tejagvd6@njit.edu}
\IEEEauthorblockA{\textsuperscript{3}George Mason University, Fairfax, USA; \quad Email-avurity@gmu.edu}
\IEEEauthorblockA{\textsuperscript{4}University of Southern Queensland, Queensland, Australia; \quad Email- Tazeem.Ahmad@unisq.edu.au}
\IEEEauthorblockA{\textsuperscript{5}University of Aveiro, Aveiro, Portugal; \quad Email- m.adnan1821@gmail.com}
\IEEEauthorblockA{\textsuperscript{6}Edinburgh Napier University, Edinburgh; \quad Email- s.rafi@napier.ac.uk}
\IEEEauthorblockA{\textsuperscript{7}Northwood High School, Irvine, CA, USA; \quad Email- patrosaisha@gmail.com, sainapatro@gmail.com}
}

% \author{
% \IEEEauthorblockN{ Tejaswani Dash}
% \IEEEauthorblockA{\textit{Capital Group} \\
% Irvine, CA, USA\\
% tejaswani.dash@capgroup.com}\and
% \IEEEauthorblockN{ Gautam Datla }
% \IEEEauthorblockA{\textit{NJIT} \\
% Newark, New Jersey, U.S.A \\
% gvd6@njit.edu}
% \and
% \IEEEauthorblockN{ Anudeep Vurity}
% \IEEEauthorblockA{\textit{George Mason University} \\
% Fairfax, U.S.A \\
% avurity@gmu.edu} 
% \and
% \IEEEauthorblockN{Tazeem Ahmad}
% \IEEEauthorblockA{\textit{University of Southern Queensland} \\
% Queensland, Australia \\ 
% Tazeem.Ahmad@unisq.edu.au}

% \and
% \IEEEauthorblockN{Mohd Adnan}
% \IEEEauthorblockA{\textit{University of Aveiro} \\
% Aveiro, Pourtugal \\ 
% m.adnan1821@gmail.com}
% }

\maketitle

\begin{abstract}
Cardiovascular disease (CVD) remains the leading cause of mortality worldwide, underscoring the need for reliable and efficient predictive tools that can support early intervention. Traditional diagnostic approaches often rely on handcrafted features and clinician expertise, while machine learning methods improve reproducibility but struggle with generalization across noisy, heterogeneous data. In this work, we propose Residual GRU+MHSA, a compact deep learning architecture tailored to tabular clinical records. The model combines residual bidirectional gated recurrent units (BiGRUs) for sequential modeling of feature columns, channel reweighting (CR) block, and multi-head self-attention (MHSA) pooling with a learnable classification token to capture global context. For a small-scale benchmark study, we evaluate the model on the UCI Heart-Disease dataset using 5-fold stratified cross-validation, benchmarking against classical methods (e.g., Logistic Regression, Random Forest, SVM) and modern deep baselines (DeepMLP, CNNs, LSTMs, Transformers). Residual GRU+MHSA achieves 0.861 accuracy, 0.860 macro-F1, 0.908 ROC-AUC, and 0.904 PR-AUC, outperforming baselines. An ablation study confirms the contributions of residual recurrence, channel gating, and attention pooling. t-SNE visualizations further show that the model learns compact, discriminative embeddings that separate disease and non-disease classes more effectively than raw features. These findings suggest that lightweight hybrid recurrent–attention architectures provide a strong balance between accuracy and efficiency for clinical risk prediction, offering a pathway toward practical deployment in resource-constrained healthcare settings.
\end{abstract}

\begin{IEEEkeywords}
Cardiovascular disease (CVD), Residual GRU, Multi-Head Self-Attention (MHSA)
\end{IEEEkeywords}

\section{Introduction}
\label{sec:intro}

As one of the most significant global health threats, cardiovascular disease (CVD) has become the leading cause of mortality worldwide. According to statistics from the World Health Organization, CVD is responsible for millions of deaths annually, exerting not only a severe impact on patient quality of life but also imposing a substantial economic burden on healthcare systems and society at large \cite{WHO2024}. Early prediction and timely intervention are therefore critical to reducing morbidity and mortality. With the rapid progress of medical technology, particularly in sensor-based data collection and computational analytics, the diagnosis and prevention of CVD have increasingly shifted from manual clinical judgment to data-driven methodologies \cite{Hossain2024}. This shift creates new possibilities for personalized healthcare and preventive strategies.

Traditional diagnostic methods for heart disease often relied on the subjective experience and judgment of clinicians, making them susceptible to variability and limited reproducibility. In contrast, machine learning (ML) methods have shown significant potential in detecting complex patterns within patient records, thereby improving predictive accuracy \cite{Doe2024}. By analyzing multiple risk factors, such as age, blood pressure, cholesterol levels, and electrocardiogram (ECG) features, ML models are able to assess an individual’s likelihood of developing CVD with greater objectivity and precision. Previous studies have demonstrated the utility of ML across healthcare tasks ranging from skin cancer classification  to cardiovascular risk stratification, highlighting the versatility of these approaches \cite{Hussein2024}. The emergence of deep learning has further strengthened predictive capabilities by automating feature extraction and learning hierarchical representations directly from raw signals. Models such as recurrent neural networks (RNNs) and long short-term memory (LSTM) networks capture sequential dependencies in time-series data, making them well-suited for heart rate forecasting and continuous health monitoring \cite{Rao2023,Vage2023}. Attention-based mechanisms and Transformer architectures extend these benefits by modeling long-range temporal correlations and providing context-aware predictions \cite{Yan2025,Yu2024}. Moreover, explainable artificial intelligence (XAI) frameworks, such as AIX360 , enable clinicians to interpret model outputs, improving trust and adoption in practice. These advances have expanded predictive accuracy and interpretability of computational tools for CVD management \cite{Arya2021},\cite{Patro2023}.

Beyond accuracy improvements, ML also offers practical advantages in healthcare delivery. Automated models process data at scale and in real time, enabling faster diagnoses and more immediate interventions. This adaptability allows models to continuously update with new data, refining predictions as medical environments evolve and new risk factors emerge. Such capabilities make ML-driven prediction a valuable tool not only for individual patient care but also for public health management and resource allocation \cite{Sengupta2019}. Despite these promising developments, challenges remain in ensuring robustness across demographic groups, reducing computational overhead for wearable deployment, and standardizing evaluation protocols, motivating further research into lightweight yet accurate architectures for CVD prediction. 

Motivated by these challenges, our study is guided by two key research questions. First, we ask whether hybrid deep learning architectures that combine recurrent sequence modeling with attention-based pooling can improve predictive performance and generalization for cardiovascular disease (CVD) risk prediction across heterogeneous populations (\textbf{RQ1}). 
Second, we investigate whether architectural design choices such as residual recurrence, attention pooling, and channel reweighting can reduce computational cost while maintaining accuracy, thereby supporting deployment in resource-constrained environments such as wearable health monitors (\textbf{RQ2}). 
These questions frame the methodological design and experimental evaluation presented in this work.

To address these challenges, this paper makes the following contributions:
\begin{itemize}
    \item We propose \textbf{Residual GRU+MHSA}, a novel deep learning model that combines bidirectional GRUs with residual connections, channel reweighting, and multi-head self-attention pooling. This design captures sequential dependencies across features, highlights informative channels, and extracts global context in tabular clinical data.
    \item We benchmark the proposed model against a wide set of classical ML methods and strong deep learning baselines under a unified 5-fold stratified cross-validation protocol, showing that Residual GRU+MHSA achieves the best overall balance across Accuracy, Macro-F1, ROC-AUC, and PR-AUC.
    \item We conduct a detailed ablation study to quantify the contribution of each component (residual recurrence, MHSA pooling, SE gating, feature dropout, and hidden dimensionality), confirming that the integrated design leads to improved robustness and generalization.
    \item We visualize learned representations using t-SNE, showing that our model transforms noisy tabular inputs into a structured latent space with clear class separation, thereby enhancing interpretability for clinical prediction tasks.
\end{itemize}

\textbf{Paper organization.} The remainder of this paper is structured as follows. Section~\ref{sec:related} reviews related work in statistical, ML, and deep learning approaches for heart rate prediction. Section~\ref{methodology} introduces the proposed hybrid LSTM-Transformer framework. Section~\ref{results} presents experimental results and ablation studies. Section~\ref{discussion} discusses practical implications and limitations. Finally, Section~\ref{sec:conclusion} concludes the paper and outlines directions for future work.

\section{Related Work}
\label{sec:related}
\begin{figure*}[ht]
    \centering
    \includegraphics[width=1\linewidth]{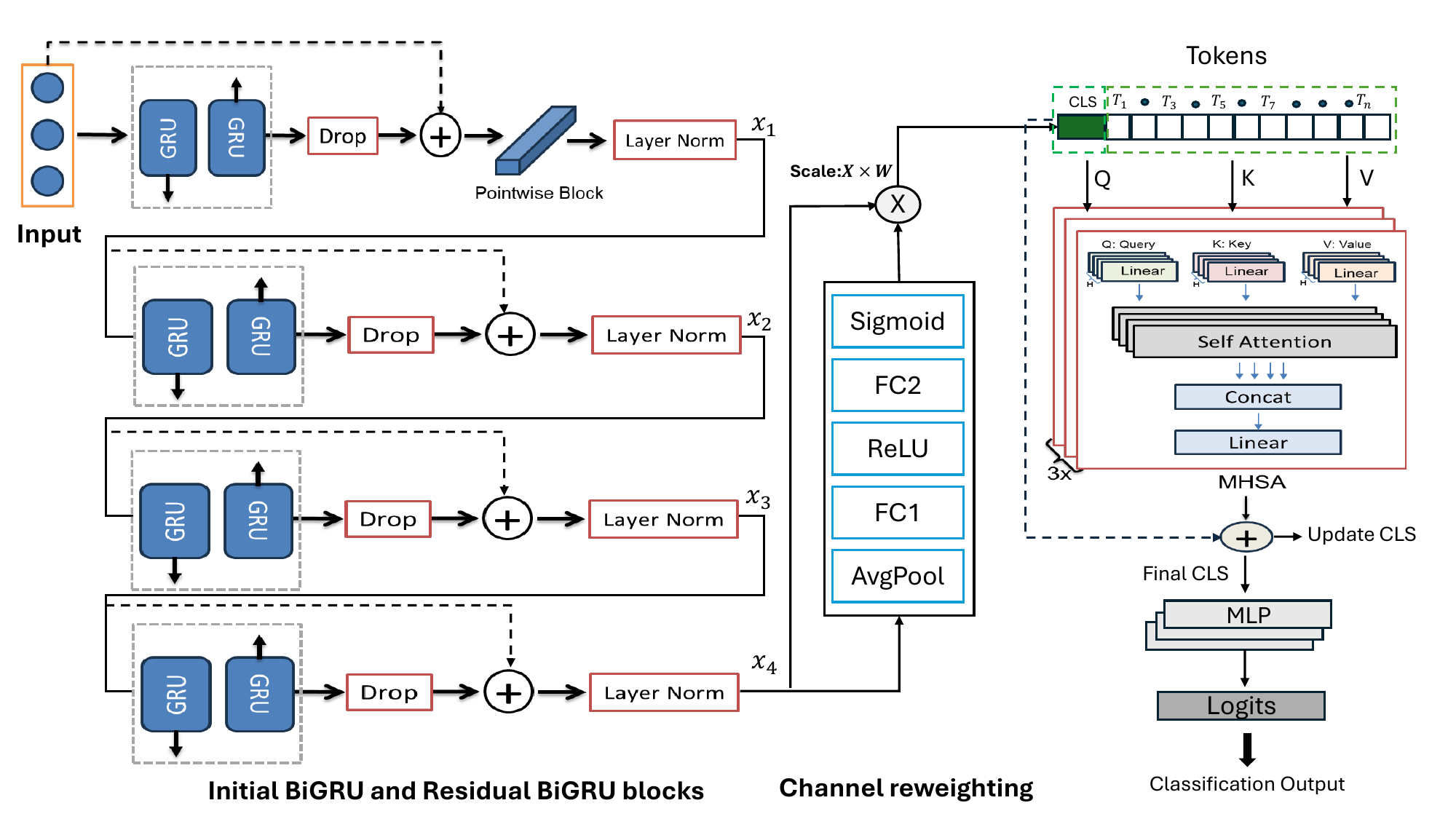}
    \caption{Overall architecture of the proposed Residual BiGRU–MHSA model.}
    \label{fig1}
\end{figure*}
Early research in heart rate and heart disease prediction relied heavily on statistical forecasting models. Linear regression and autoregressive integrated moving average (ARIMA) models were frequently adopted to capture temporal dynamics in physiological data \cite{Do2022}. These approaches offered interpretability and low computational cost, but they were limited in representing nonlinear physiological patterns and often underperformed in heterogeneous populations or under physical stress conditions. In particular, age-specific variability and exercise-induced fluctuations exposed the fragility of purely statistical formulations.

The shift toward ML introduced models that could capture more complex interactions between features. Algorithms such as support vector regression (SVR), random forests, and gradient boosting demonstrated improved accuracy by accounting for nonlinear dependencies \cite{Vijaya2022}. Comparative studies revealed that these models consistently outperformed classical baselines, though their effectiveness still varied with demographic subgroup and data availability. For instance, prediction errors were systematically higher for elderly cohorts, indicating that generalized models struggled to adapt to age-specific physiology.

Recurrent architectures, particularly RNNs and long short-term memory (LSTM) networks, marked a turning point for continuous health monitoring. Their ability to retain temporal dependencies made them effective for heart rate estimation and disease risk modeling. Rao \textit{et al.}\ \cite{Rao2023} integrated RNN and LSTM within an edge computing framework for wearable monitoring systems, enabling low-latency, resource-efficient inference. Similarly, Våge \textit{et al.}  compared LSTM against tree-based ensembles (XGBoost, LightGBM), reporting superior temporal accuracy for LSTM in dynamic monitoring settings \cite{Vage2023}. Bian \textit{et al.} extended this line by proposing privacy-preserving LSTM frameworks, ensuring secure estimation of heart rate in distributed IoT environments \cite{Bian2023} . Recent work has also extended token-based sequence modeling beyond healthcare, for example by framing typhoon tracks as token sequences and applying large language model-empowered adversarial fusion for improved prediction \cite{luo2025large}.

Recent work emphasizes attention mechanisms and hybrid deep learning models to capture richer contextual information. Yan \textit{et al.} \cite{Yan2025} proposed an enhanced spatio-temporal attention mechanism, showing that context-aware models can better model sequential dependencies in biomedical signals. Yu \textit{et al. }\cite{Yu2024} highlighted the role of large language models (LLMs) in multimodal learning, suggesting future integration of textual, signal, and demographic data streams for robust cardiovascular modeling. These advances point toward architectures capable of combining physiological signals with contextual features such as lifestyle or environmental variables. \textit{Yu et al.}\cite{Yu2024TransformerPSO} optimized a Transformer-based heart disease prediction model using particle swarm optimization, reporting improved accuracy over traditional ensemble methods.

Despite progress, several limitations persist. First, model performance often degrades in elderly or high-intensity exercise populations, where physiological variability is high \cite{Do2022,Vijaya2022}. Second, computational complexity remains a barrier to real-time deployment in wearable or mobile systems, motivating lightweight yet accurate architectures \cite{Rao2023}. Finally, the absence of standardized benchmarking datasets and evaluation protocols hinders reproducibility. Many reported studies rely on different subsets, preprocessing pipelines, or non-comparable metrics, complicating direct performance comparisons. These gaps highlight the need for frameworks that jointly address accuracy, efficiency, and adaptability while grounding evaluation in standardized, demographically diverse cohorts.

Recent advances in sequential modeling have demonstrated the effectiveness of combining recurrent architectures with attention and domain-specific enhancements across diverse engineering tasks. Bai and Tahmasebi \cite{Bai2022} introduced an attention-based LSTM-FCN for earthquake detection and location, showing that attention mechanisms can enhance feature extraction from sequential signals. Similarly, Zhang et al.\ \cite{Zhang2023SSA} applied a singular spectrum analysis (SSA) integrated with LSTM for intelligent real-time prediction of multi-region thrust in shield machines, achieving robust performance in geotechnical engineering applications. In tunneling, Zhang and Zhao \cite{Zhang2023MCNN} proposed a novel wear rate index alongside a multi-channel CNN-GRU model for real-time cutter wear prediction, highlighting the benefits of hybrid recurrent–convolutional designs. Beyond civil engineering, Yao \textit{et al. }\cite{Yao2023} developed a physics-informed, multi-step GRU-based framework for conflict-based vehicle safety prediction, integrating domain knowledge with deep sequential modeling. Collectively, these studies underscore the growing adoption of hybrid and attention-enhanced architectures in safety-critical, real-time prediction tasks, motivating their adaptation to biomedical signal forecasting.

\section{Methodology} \label{methodology}

Our Residual GRU + MHSA framework integrates recurrent modeling, feature reweighting, and attention pooling into a compact sequence classifier for tabular data. Each input feature is embedded into a higher-dimensional space, with the $d$ columns treated as a sequence of tokens. An initial bidirectional GRU captures short-range dependencies, while a stack of residual BiGRU blocks extends temporal modeling and maintains stable gradients through skip connections. A channel reweighting block then adaptively scales feature dimensions based on global context. The reweighted sequence is aggregated using a learnable class token (CLS), which attends to all features through multi-head self-attention (MHSA) pooling to capture long-range interactions. The final CLS representation is passed through a multi-layer perceptron with dropout to produce class probabilities, optimized under a class-weighted and label-smoothed loss. This design leverages sequential structure, adaptive channel scaling, and attention-based pooling to achieve robust predictions.

\subsection{Problem formulation}
Given $(x_i,y_i)_{i=1}^n$, we treat each feature vector $x_i \in \mathbb{R}^d$ as a sequence of $d$ tokens, each corresponding to one input column. The model outputs a probability $p_\theta(y=1 \mid x) \in [0,1]$. Training uses class-weighted binary cross-entropy with label smoothing $\varepsilon$:
\begin{align}
\tilde y_i &= 
\begin{cases}
1-\varepsilon, & y_i=1,\\
\varepsilon,   & y_i=0,
\end{cases}\\
\mathcal{L}(\theta) &= -\sum_{i=1}^n w_{y_i}\Big[\tilde y_i \log p_\theta(x_i) + (1-\tilde y_i)\log(1-p_\theta(x_i))\Big]
\end{align}
where $w_c$ are inverse-frequency class weights normalized to the number of classes, and $\varepsilon$ follows \cite{Muller2019LabelSmoothing}.

\subsection{Proposed model: Residual GRU+MHSA}
We propose a compact sequence model tailored to tabular columns: a residual BiGRU stack with multi-head self-attention pooling (MHSA) and channel reweighting (CR). The BiGRU layers capture local dependencies across features while residual connections stabilize training and enable deeper recurrent modeling. Channel reweighting adaptively emphasizes informative feature dimensions, reducing noise from less relevant inputs. Finally, a learnable CLS token with MHSA aggregates the sequence into a global representation, allowing the model to capture long-range dependencies before classification through a lightweight feedforward head, see Fig.\ref{fig1}.

\paragraph{\textbf{Input embedding}}
We treat the $d$ columns of a tabular sample as a length-$d$ sequence, 
where each scalar feature is embedded into a $d_{\text{model}}$-dimensional vector. 
This embedding step places heterogeneous tabular features into a shared representation space, 
making them amenable to sequential modeling. To encourage robustness and reduce overfitting, 
we apply column-wise feature dropout during training, randomly masking individual feature embeddings 
with probability $p_f$.

\begin{equation}
E=\operatorname{Linear}(x)\in\mathbb{R}^{d\times d_{\mathrm{model}}}.
\label{eq:embed}
\end{equation}

We apply feature dropout (column-wise) with rate $p_f$ during training.

\paragraph{\textbf{Initial BiGRU and residual BiGRU blocks}}

An initial bidirectional GRU transforms the embeddings into a feature space of dimension $f$, capturing short-range dependencies across neighboring features in both directions of the sequence. 
To further improve model capacity, we stack $N$ residual BiGRU blocks, inspired by the residual design used in 
vision-based architectures such as ColFigPhotoAttnNet \cite{vurity2025colfigphotoattnnet}. Each block applies a BiGRU transformation followed by dropout and adds the result back to its input before layer normalization. 
This residual design stabilizes training, preserves earlier representations, and mitigates gradient degradation, allowing the recurrent stack to model longer-range dependencies without sacrificing 
convergence. In our experiments, we set $N{=}3$, resulting in a total of four BiGRU layers. 

An initial BiGRU maps the embedded sequence $E$ to $H^{(0)} \in \mathbb{R}^{d \times f}$, 
where the output dimension is
\begin{equation}
f =
\begin{cases}
2 d_{\text{model}}, & \text{if bidirectional}, \\[4pt]
d_{\text{model}}, & \text{otherwise}.
\end{cases}
\end{equation}
We add a residual projection from the input embedding and apply layer normalization:
\begin{subequations}\label{eq:resbigru}
\setlength{\abovedisplayskip}{4pt}\setlength{\belowdisplayskip}{4pt}
\begin{align}
H^{(0)} &= \operatorname{LN}\!\big(\operatorname{BiGRU}(E) + \operatorname{Proj}(E)\big), \\
\intertext{and then stack $N$ residual BiGRU blocks, each of the form:}
H^{(\ell)} &= \operatorname{LN}\!\big(H^{(\ell-1)} + \operatorname{BiGRU}(H^{(\ell-1)})\big), 
\quad \ell = 1, \dots, N.
\end{align}
\end{subequations}

\begin{figure}
    \centering
    \includegraphics[width=0.82\linewidth,height=0.25\textheight]{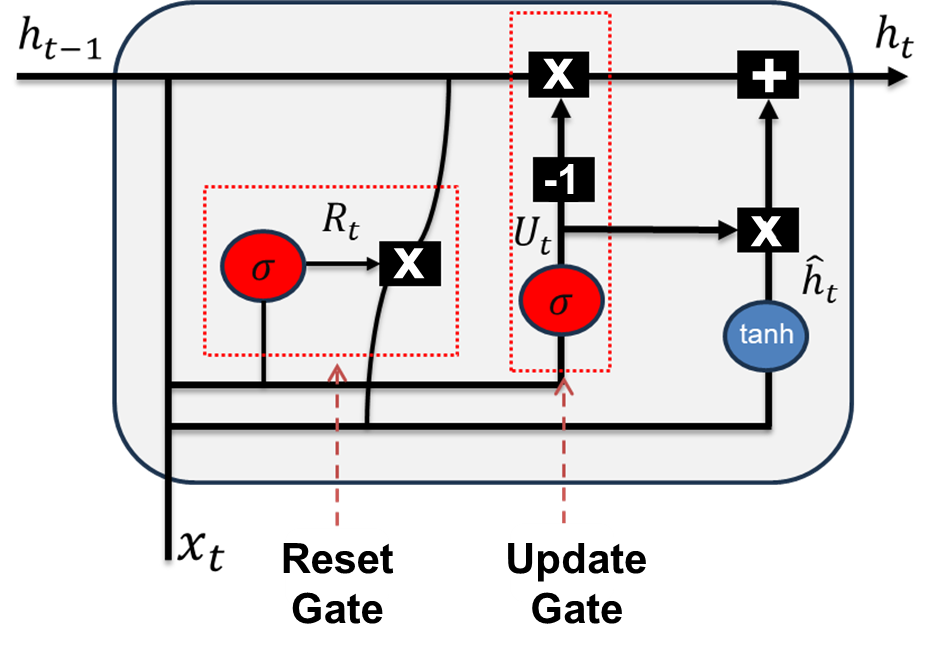}
    \caption{Architecture of a gated recurrent unit (GRU). 
    The reset gate $R_t$ controls how much of the past state $h_{t-1}$ contributes to the candidate hidden state $\hat{h}_t$, 
    while the update gate $U_t$ balances information from $h_{t-1}$ and $\hat{h}_t$ to form the new state $h_t$.}
    \label{fig2}
\end{figure}

RNNs are powerful for modeling sequential dependencies 
but often face optimization difficulties such as vanishing and exploding gradients \cite{elman1990rnn}. 
Long short-term memory (LSTM) units~\cite{hochreiter1997lstm} address this by introducing explicit memory cells and gating mechanisms. 
Gated recurrent units (GRUs)~\cite{cho2014gru} simplify this architecture with reset and update gates, 
which regulate how information is updated and forgotten at each timestep. 
Specifically, the reset gate $R_t$ determines how much of the past hidden state should be ignored when computing the candidate hidden state $\hat{h}_t$, 
while the update gate $U_t$ interpolates between the previous hidden state $h_{t-1}$ and the candidate $\hat{h}_t$ to produce the updated hidden state $h_t$. 
Fig.\ref{fig2} design balances efficiency and expressiveness, requiring fewer parameters than LSTMs while still effectively capturing both short- and long-range dependencies.

\paragraph{\textbf{Channel reweighting, MHSA pooling with CLS, and prediction head}}
To aggregate the reweighted sequence into a compact representation, 
we introduce a learnable classification token (CLS). 
At each self-attention layer, the CLS embedding is first normalized and projected into a query vector, 
while the hidden sequence is normalized and projected into keys and values. 
Through scaled dot-product attention, the CLS token attends to all feature tokens and integrates their information. 
The output is added residually to the previous CLS state, producing an updated sequence-level summary. 
Repeating this process for $L$ layers allows the CLS token to iteratively refine its representation 
and capture long-range dependencies that may be missed by recurrent modeling alone. Other work in the field of EEG has also shown  benefit from learning compact temporal–channel representations \cite{faye2024k, abdullah2025hypergcn}.

After the final attention layer, the CLS embedding serves as a global representation of the input. 
We normalize this embedding and pass it through a multi-layer perceptron with dropout and nonlinear activations. 
The MLP projects the sequence-level representation into a lower-dimensional latent space before producing a final scalar output. 
A sigmoid activation converts this score into a probability for binary classification. 
Training is performed with class-weighted and label-smoothed cross-entropy loss, 
which improves robustness under class imbalance and yields better-calibrated predictions.

We begin with the BiGRU output sequence $H^{(N)} \in \mathbb{R}^{T \times f}$,
where $T$ is the sequence length and $f$ is the feature dimension.  
To highlight informative channels, First, we compute a global summary vector by temporal average pooling:
\begin{align}
s = \tfrac{1}{T} \sum_{t=1}^T H^{(N)}_t .
\end{align}
This summary is projected through two fully connected layers with
nonlinearity $\phi(\cdot)$ and a sigmoid gate:
\begin{align}
w = \sigma\!\big(W_2 \phi(W_1 s)\big) 
\end{align}
where $w \in \mathbb{R}^f$ acts as a per-channel importance weight.  
Each timestep is then reweighted elementwise:
\begin{align}
\hat H_t = H^{(N)}_t \odot w 
\end{align}

To pool the sequence into a compact representation, we use a CLS.  
At each Transformer layer $l$, the CLS state of the previous layer
is normalized and linearly projected to form the query:
\begin{align}
Q^{(l)} = \operatorname{LN}(\mathrm{CLS}^{(l-1)}) W_Q 
\end{align}
Similarly, the reweighted hidden sequence is normalized and projected
to obtain keys and values:
\begin{align}
K^{(l)} = \operatorname{LN}(\hat H) W_K, 
\qquad
V^{(l)} = \operatorname{LN}(\hat H) W_V 
\end{align}

The CLS token attends to the reweighted sequence via scaled
dot-product attention:
\begin{align}
A^{(l)} = \operatorname{softmax}\!\left(\frac{Q^{(l)} {K^{(l)}}^\top}{\sqrt{d_k}}\right) 
\end{align}
The attended information is aggregated to update the CLS representation:
\begin{align}
\mathrm{CLS}^{(l)} = \mathrm{CLS}^{(l-1)} + A^{(l)} V^{(l)} 
\end{align}
This process repeats for $l = 1, \dots, L$ layers.

After the final layer, the CLS token provides a sequence-level embedding.  
We normalize it and map it through two fully connected layers with nonlinearity:
\begin{align}
z &= \operatorname{LN}(\mathrm{CLS}^{(L)}) \\
o &= W_2 \phi(W_1 z) .
\end{align}
Finally, a sigmoid outputs the probability for the binary prediction task:
\begin{align}
p_\theta(y{=}1 \mid x) = \sigma(o) 
\end{align}

In all experiments we use $N{=}3$ residual BiGRU blocks, $L{=}3$ MHSA layers, $d_{\mathrm{model}}{=}128$, dropout $p{=}0.2$, and feature dropout $p_f{=}0.1$, chosen via small cross-validated ablations on the training folds to maximize macro-F1 and stability; these hyperparameters are then fixed for all runs.

\subsection{Training and optimization}
We use AdamW \cite{LoshchilovHutter2019AdamW} with learning rate $8\cdot 10^{-4}$, weight decay $6\cdot 10^{-5}$, cosine-annealing schedule (SGDR) over the epoch budget \cite{LoshchilovHutter2017SGDR}, gradient clipping at $5.0$, label smoothing $\varepsilon=0.05$, and early stopping on macro-F1 with patience $18$. For class imbalance, we apply inverse-frequency class weights computed on the training split of each fold. Batch size is $32$.

\section{Experimental setup and results} \label{results}

\subsection{Dataset and Preprocessing}
We use a small but heavily used benchmark dataset called the UCI Heart Disease (Cleveland) dataset (\textit{id}=45) from the UCI ML Repository \cite{janosi1989heartdisease}. After one-hot encoding categorical fields and median-imputing missing values, the feature matrix $X\in\mathbb{R}^{n\times d}$ (with $d$ columns after encoding) and binary target $y\in\{0,1\}^n$ are formed by mapping \texttt{num>0} to $1$ and \texttt{num=0} to $0$. The aim  is to perform a binary classification in order to predict the presence of heart disease. All pre-processing is performed \emph{inside each training fold} to avoid leakage. In  order to match common tabular ML practice, in this pre-processing we apply one-hot encoding of categorical variables, median imputation, and standardization using only the training fold statistics \cite{pedregosa2011scikit}.

\subsection{Baselines}
% We report classical baselines from scikit-learn \cite{pedregosa2011scikit}: Logistic Regression, GaussianNB, SVM (RBF) \cite{cortes1995support}, Extra Trees \cite{geurts2006extremely}, Random Forest \cite{breiman2001random}, kNN \cite{cover1967nearest}, Gradient Boosting \cite{friedman2001greedy}, and CART \cite{breiman1984classification}. Deep learning baselines are: a DeepMLP \cite{goodfellow2016deep}, a 1D CNN \cite{lecun1998gradient}, a stacked LSTM \cite{hochreiter1997long}, a stacked GRU \cite{cho2014learning}, and a Transformer encoder \cite{vaswani2017attention}. All deep models are implemented in PyTorch with identical CV protocol, per-fold scaling, early stopping, and metric computation.

To provide a comprehensive benchmark, we evaluate both classical and deep learning baselines widely recognized in the literature. The classical models are implemented using \texttt{scikit-learn} \cite{pedregosa2011scikit} and include Logistic Regression, Gaussian Naive Bayes (GaussianNB), Support Vector Machine with a Radial Basis Function kernel (SVM-RBF) \cite{cortes1995support}, Extra Trees \cite{geurts2006extremely}, Random Forest \cite{breiman2001random}, k-Nearest Neighbours (kNN) \cite{cover1967nearest}, Gradient Boosting \cite{friedman2001greedy}, and Classification and Regression Trees (CART) \cite{breiman1984classification}. These models serve as interpretable yet competitive baselines for small- to medium-scale datasets and allow comparison across linear, probabilistic, and ensemble-based paradigms.

For deep learning baselines, we employ architectures representative of current sequence and representation learning approaches, a fully connected Deep Multi-Layer Perceptron (DeepMLP) \cite{goodfellow2016deep}, a one-dimensional Convolutional Neural Network (1D-CNN) \cite{lecun1998gradient}, a stacked Long Short-Term Memory network (LSTM) \cite{hochreiter1997long}, a stacked Gated Recurrent Unit model (GRU) \cite{cho2014learning}, and a Transformer encoder \cite{vaswani2017attention}. These models capture progressively complex temporal and contextual dependencies, enabling assessment of how network depth and architectural design influence performance on our task.

All deep models are implemented in \texttt{PyTorch} under a unified cross-validation (CV) protocol. Feature scaling and normalization are applied per fold to prevent data leakage. Early stopping is used to mitigate overfitting, and all metrics are computed on held-out test folds to ensure comparability and robustness across baselines.

\subsection{Evaluation protocol}
The performance of the model is assessed using a 5-fold stratified cross-validation \cite{kohavi1995study}, ensuring that the distribution of the labels is preserved in the different sections. In each fold, we first fit all preprocessing steps (one-hot encoding, imputation, and standardization) exclusively on the training split to prevent information leakage then we train the model with early stopping \cite{prechelt1998early} based on validation loss to avoid overfitting, and finally we evaluate using four complementary metrics i.e., Accuracy, macro-F1, ROC-AUC, and PR-AUC \cite{davis2006relationship,sokolova2009systematic}. Because threshold-dependent metrics (Accuracy and F1) can vary with the decision boundary, we tune a scalar probability threshold $\tau$ on the validation split by sweeping $\tau \in [0.1, 0.9]$ in increments of 0.01 and selecting $\tau^{\star}$ that maximizes macro-F1. Final results are reported as the mean $\pm$ standard deviation across folds, while ROC-AUC and PR-AUC are reported.

\begin{table*}[htbp]
\centering
\caption{Baseline results on the UCI Heart Disease dataset (5-fold stratified CV).
Reported are mean $\pm$ std for Accuracy, Macro-F1, ROC-AUC, and PR-AUC. Best results are highlighted in \textbf{bold}.}
\label{tab:all_baselines}
\tiny
\resizebox{0.95\linewidth}{!}{%
\begin{tabular}{lcccc}
\hline
\textbf{Model} & \textbf{Accuracy} & \textbf{F1 (Macro)} & \textbf{ROC-AUC} & \textbf{PR-AUC} \\
\hline
\multicolumn{5}{c}{\textit{Classical baselines}} \\
\hline
Logistic Regression & $0.832 \pm 0.050$ & $0.830 \pm 0.050$ & $0.912 \pm 0.018$ & $0.908 \pm 0.035$ \\
GaussianNB          & $0.855 \pm 0.040$ & $0.853 \pm 0.040$ & $0.902 \pm 0.023$ & $0.895 \pm 0.032$ \\
SVM (RBF)           & $0.838 \pm 0.039$ & $0.835 \pm 0.040$ & $0.898 \pm 0.025$ & $0.900 \pm 0.040$ \\
Extra Trees         & $0.822 \pm 0.028$ & $0.820 \pm 0.028$ & $0.915 \pm 0.025$ & $0.895 \pm 0.048$ \\
Random Forest       & $0.822 \pm 0.034$ & $0.819 \pm 0.034$ & $0.910 \pm 0.023$ & $0.907 \pm 0.029$ \\
kNN                 & $0.818 \pm 0.019$ & $0.816 \pm 0.020$ & $0.903 \pm 0.029$ & $0.880 \pm 0.043$ \\
Gradient Boosting   & $0.802 \pm 0.048$ & $0.799 \pm 0.048$ & $0.886 \pm 0.037$ & $0.876 \pm 0.041$ \\
Decision Tree       & $0.765 \pm 0.047$ & $0.764 \pm 0.046$ & $0.766 \pm 0.043$ & $0.676 \pm 0.060$ \\
\hline
\multicolumn{5}{c}{\textit{Deep learning baselines}} \\
\hline
DeepMLP         & $0.855 \pm 0.037$ & $0.852 \pm 0.038$ & $0.902 \pm 0.019$ & $0.907 \pm 0.029$ \\
DeepCNN1D       & $0.835 \pm 0.039$ & $0.833 \pm 0.040$ & $0.890 \pm 0.037$ & $0.891 \pm 0.035$ \\
StackedLSTM     & $0.851 \pm 0.038$ & $0.849 \pm 0.038$ & $0.891 \pm 0.023$ & $0.893 \pm 0.034$ \\
StackedGRU      & $0.851 \pm 0.038$ & $0.848 \pm 0.039$ & $0.897 \pm 0.021$ & $0.899 \pm 0.033$ \\
DeepTransformer & $0.848 \pm 0.042$ & $0.847 \pm 0.042$ & $0.909 \pm 0.027$ & $0.912 \pm 0.031$ \\
\hline
\multicolumn{5}{c}{\textit{Hybrid deep baselines}} \\
\hline
CNN--LSTM          & $0.855 \pm 0.034$ & $0.852 \pm 0.034$ & $0.890 \pm 0.022$ & $0.892 \pm 0.028$ \\
CNN--GRU           & $0.845 \pm 0.037$ & $0.842 \pm 0.038$ & $0.888 \pm 0.025$ & $0.889 \pm 0.034$ \\
LSTM--Transformer  & $0.858 \pm 0.043$ & $0.856 \pm 0.044$ & $0.907 \pm 0.020$ & $0.907 \pm 0.020$ \\
\hline
Residual GRU+MHSA (ours) & $0.861 \pm 0.032$ & $0.860 \pm 0.032$ & $0.908 \pm 0.022$ & $0.904 \pm 0.027$ \\
\hline
\end{tabular}}
\end{table*}

\subsection{Results}

Table~\ref{tab:all_baselines} presents the results of all evaluated models on the UCI Heart Disease dataset under 5-fold stratified cross-validation. We organize the results into three baseline groups, classical machine learning, deep learning, and hybrid architectures, followed by our proposed Residual GRU+MHSA model. Each method is evaluated using Accuracy, macro-F1, ROC-AUC, and PR-AUC, with values reported as mean $\pm$ standard deviation across folds. This allows us to compare both predictive power and stability across different model classes.

In the \textit{classical baseline} group, GaussianNB achieves the highest accuracy ($0.855$) and macro-F1 ($0.853$), suggesting that simple probabilistic assumptions still capture meaningful structure in the dataset. Extra Trees records the strongest ROC-AUC ($0.915$), reflecting its ability to rank predictions well despite modest classification accuracy. Logistic Regression and Random Forest remain competitive and stable across folds, consistent with their wide use in medical risk prediction tasks. On the other hand, single Decision Trees underperform substantially, achieving only $0.765$ accuracy and $0.766$ ROC-AUC, underscoring the need for ensemble or regularization strategies to improve generalization.

Moving to the \textit{deep learning baselines}, we observe a more balanced performance profile. DeepMLP matches GaussianNB in accuracy ($0.855$) while reducing variance across folds, showing that even fully connected networks can robustly capture tabular patterns when carefully regularized. Recurrent models such as StackedLSTM and StackedGRU further enhance sequential dependency modeling, with StackedGRU achieving a ROC-AUC of $0.897$ and PR-AUC of $0.899$, outperforming purely feed-forward baselines. DeepTransformer stands out with the strongest ranking performance (ROC-AUC $0.909$, PR-AUC $0.912$), validating the strength of attention-based architectures in capturing non-local dependencies across features. Notably, all deep models substantially outperform weaker classical methods such as CART or kNN, highlighting their capacity to learn higher-level feature interactions.

The \textit{hybrid baselines} illustrate the benefits of combining inductive biases. CNN–LSTM and CNN–GRU incorporate local feature extraction followed by sequential modeling, offering competitive results with smoother variance across folds. The LSTM–Transformer hybrid achieves the strongest accuracy ($0.858$) and macro-F1 ($0.856$) within this group, indicating that recurrent–attention combinations can leverage complementary strengths. Interestingly, while these models often improve accuracy and F1, their ROC-AUC remains slightly lower than the best Transformer, suggesting that ranking precision still favors pure attention mechanisms.

Finally, our proposed \textit{Residual GRU+MHSA} model achieves the best overall balance across all metrics: $0.861$ accuracy, $0.860$ macro-F1, $0.908$ ROC-AUC, and $0.904$ PR-AUC. This consistency reflects the architectural contributions of three key design choices. First, residual BiGRU blocks allow for deeper recurrent modeling without gradient degradation, capturing both local and global dependencies between features. Second, Channel Reweighting (CR) block adaptively reweights feature channels, amplifying informative signals while suppressing noise. Third, MHSA pooling with a learnable CLS token provides a structured mechanism for global sequence summarization, complementing the recurrent backbone with non-local attention. Together, these mechanisms enable the model to outperform both classical and deep learning baselines, not by excelling in a single metric but by achieving stable and competitive results across all four. Notably, the relatively small variance in fold-to-fold performance further highlights its robustness and generalization ability. Overall, the results indicate that while classical methods such as GaussianNB and Extra Trees can still provide strong benchmarks on smaller datasets, modern deep learning architectures—particularly those that combine recurrence, attention, and channel reweighting, offer the best trade-off between accuracy, ranking ability, and stability. Our Residual GRU+MHSA model extends these trends by unifying the strengths of recurrent and attention-based designs into a compact, interpretable, and effective architecture tailored for tabular medical prediction tasks.

\subsection{Ablation}

\begin{table}[htb]
\centering
\caption{Ablation study of the proposed Residual GRU+MHSA model on the UCI Heart Disease dataset (5-fold stratified CV). 
Reported are mean $\pm$ std for Accuracy and mean for Macro-F1, ROC-AUC, and PR-AUC.}
\label{tab:ablation}
\renewcommand{\arraystretch}{1.3} % increase row height
\scriptsize
\begin{tabular}{lcccc}
\hline
\textbf{Variant} & \textbf{Accuracy} & \textbf{F1 (Macro)} & \textbf{ROC-AUC} & \textbf{PR-AUC} \\
\hline
Residual GRU+MHSA        & $0.861 \pm 0.037$ & $0.859$ & $0.904$ & $0.908$ \\ \hline
-CR                    & $0.865 \pm 0.032$ & $0.862$ & $0.909$ & $0.911$ \\
-MHSA (mean+max)       & $0.859 \pm 0.031$ & $0.851$ & $0.891$ & $0.899$ \\
-Residual stack (N=0)  & $0.855 \pm 0.048$ & $0.852$ & $0.897$ & $0.895$ \\
Uni-GRU (no bidi)      & $0.841 \pm 0.029$ & $0.849$ & $0.891$ & $0.896$ \\
MHSA layers = 1        & $0.858 \pm 0.034$ & $0.855$ & $0.898$ & $0.905$ \\
N=2 residual blocks    & $0.859 \pm 0.032$ & $0.858$ & $0.901$ & $0.905$ \\
$d_{\mathrm{model}}=96$& $0.855 \pm 0.034$ & $0.852$ & $0.899$ & $0.898$ \\
No feature dropout     & $0.848 \pm 0.038$ & $0.845$ & $0.899$ & $0.902$ \\
Shallow head           & $0.858 \pm 0.034$ & $0.856$ & $0.901$ & $0.886$ \\
\hline
\end{tabular}
\end{table}

\begin{figure}[htbp]
    \centering
    \includegraphics[width=0.98\linewidth,height=0.13\textheight]{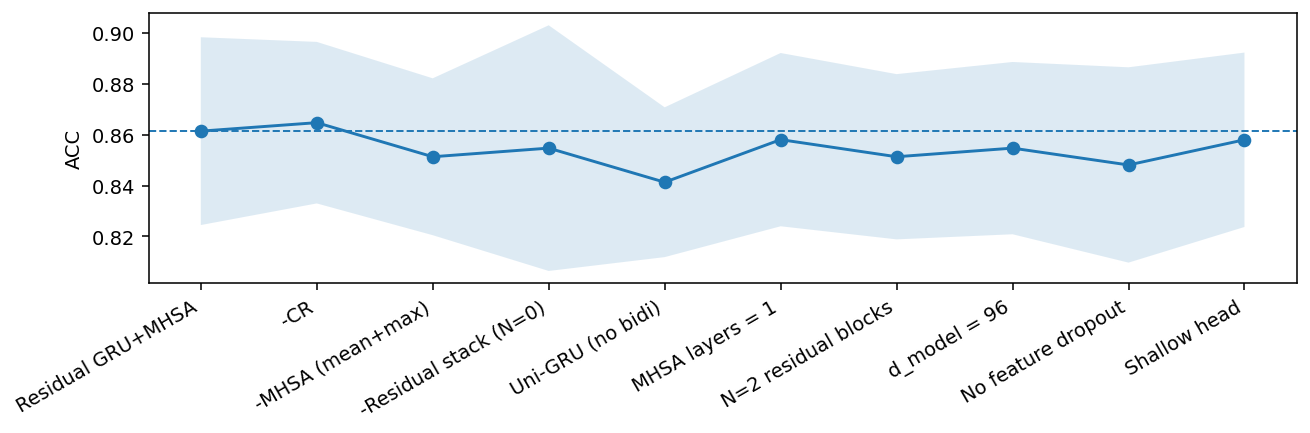}
    \caption{Ablation study of the Residual GRU+MHSA model on the UCI Heart Disease dataset. 
    The plot shows mean Accuracy across folds with standard deviation as shaded bands. 
    Each variant corresponds to the removal or modification of one component (e.g., SE gating, MHSA pooling, residual depth, bidirectionality, feature dropout). 
    This visualization highlights the relative contribution of each module to predictive performance.}
    \label{fig:ablation_acc}
\end{figure}

The ablation study  evaluates the contribution of each architectural component of Residual GRU+MHSA. To assess the contribution of each architectural component in the proposed Residual GRU+MHSA model, we conducted an ablation study, see in Table~\ref{tab:ablation}. The baseline configuration includes an input projection, an initial BiGRU layer followed by three residual BiGRU blocks,  Channel reweighting (CR), a three-layer multi-head self-attention (MHSA) pooling mechanism with a learnable CLS token, feature dropout, and a two-layer MLP prediction head. Each ablation removes or modifies one component while keeping the rest fixed, enabling a controlled analysis of its effect on predictive performance.

Removing the SE module slightly improved accuracy (0.865 vs. 0.861), suggesting that channel reweighting (CR) offers limited benefit in this relatively low-dimensional tabular domain. By contrast, replacing the MHSA pooling with simple mean–max pooling reduced ROC-AUC (0.891 vs. 0.904) and PR-AUC (0.899 vs. 0.908), confirming the importance of attention pooling for sequence-level summarization. Eliminating the residual stack (N=0) degraded accuracy to 0.855 and ROC-AUC to 0.897, showing that stacked recurrent blocks are essential for extracting hierarchical feature dependencies.

When bi-directionality was removed (Uni-GRU), accuracy dropped to 0.841, the lowest across all variants, underscoring that forward-only recurrence is insufficient and that contextual information from both directions is critical. Reducing MHSA depth to a single layer also lowered ROC-AUC (0.898 vs. 0.904), indicating that a deeper attention stack is more effective at modeling feature interactions. Using only two residual blocks achieved results close to the baseline (0.859 accuracy), but the three-block configuration remained slightly stronger, suggesting an optimal trade-off at moderate depth.

Shrinking the embedding dimension from 128 to 96 decreased accuracy to 0.855 and ROC-AUC to 0.899, reflecting a loss of representational capacity. Disabling feature dropout resulted in accuracy 0.848, highlighting its role in regularization. Finally, simplifying the head to a single projection (“shallow head”) retained accuracy (0.858) but reduced PR-AUC substantially (0.886 vs. 0.908), showing that the deeper head is especially beneficial for precision–recall trade-offs.

Overall, the ablation results reveal that bidirectionality, residual recurrence, and MHSA pooling are the most influential design choices, see Fig.\ref{fig:ablation_acc}. SE provides marginal gains at best, while feature dropout and the deeper head primarily enhance generalization and calibration. These findings validate the specific combination of components in the proposed Residual GRU+MHSA model and explain why it achieves the strongest and most balanced performance across metrics.

\subsection{t-SNE Visualization of Learned Representations}

\begin{figure}[ht]
    \centering
    \includegraphics[width=1\linewidth]{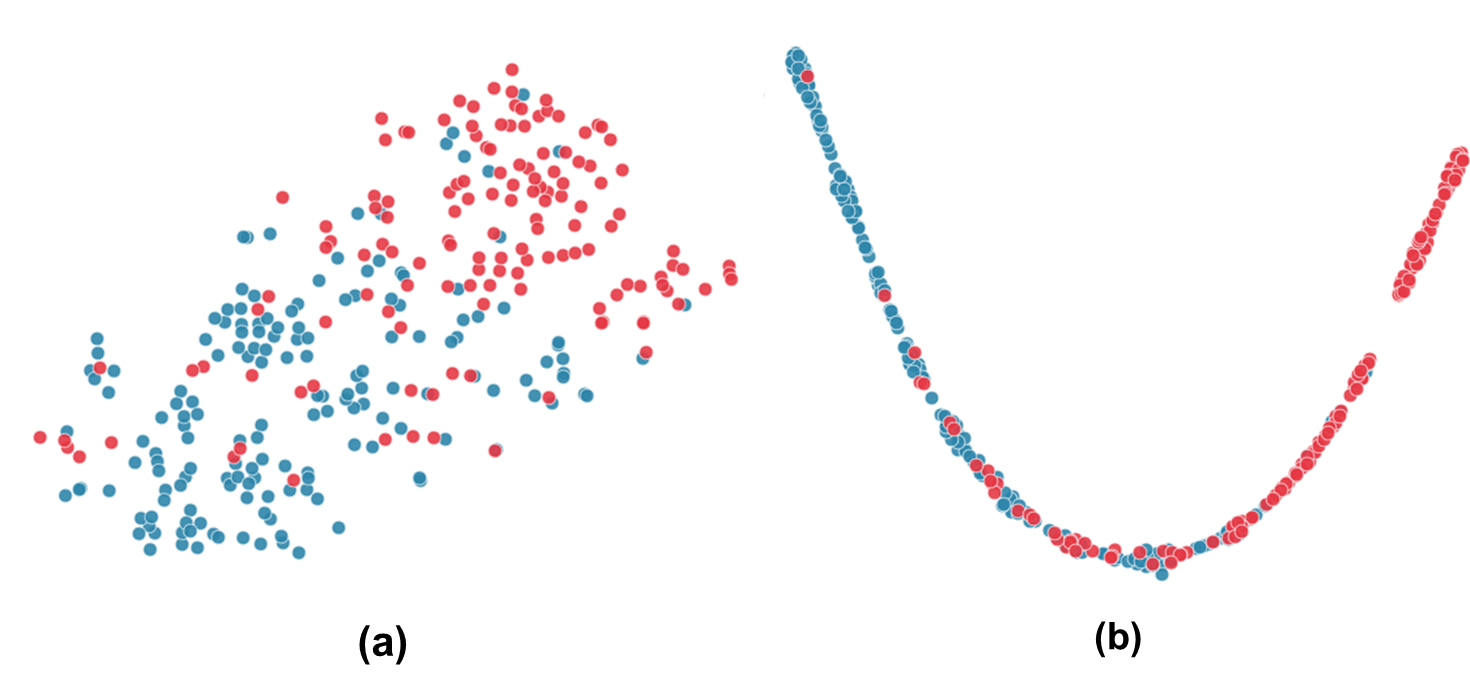}
    \caption{t-SNE visualization comparing (a) raw input features and (b) learned embeddings from the penultimate layer of the proposed model.}
    \label{tsne}
\end{figure}

To better understand the representations learned by the proposed Residual GRU+MHSA model, we apply t-Distributed Stochastic Neighbor Embedding (t-SNE) to both the raw input features and the embeddings extracted from the penultimate layer. As shown in Figure~\ref{tsne}, the raw features exhibit substantial overlap between the two classes (disease vs.\ no disease), indicating that the original feature space offers limited separability. In contrast, the learned embeddings form two distinct clusters, with samples of the same class grouped tightly together and clearer boundaries between classes. This demonstrates that the model effectively transforms noisy, high-dimensional tabular inputs into a structured latent space where decision boundaries are more discriminative. The visualization highlights how sequential modeling, channel reweighting (CR), and attention pooling jointly enhance representation learning and ultimately improve predictive performance.

\section{Discussion} \label{discussion}

The experimental results across baselines and ablation variants provide several insights into the design and performance of deep sequence models for tabular cardiovascular prediction. First, the baseline comparisons (Table~\ref{tab:all_baselines}) highlight the relative strengths of different families of models. Classical machine learning methods such as Logistic Regression, Random Forest, and Extra Trees deliver strong ROC-AUC values, consistent with their ability to exploit decision boundaries and ensemble diversity in low-dimensional tabular settings. However, their macro-F1 and overall accuracy remain limited compared to neural models, particularly in imbalanced cases where sensitivity and specificity must be balanced. Simpler learners such as single decision trees consistently underperform, reaffirming that robust generalization requires models capable of capturing nonlinear interactions among risk factors.
Among deep learning baselines, the performance gap is more nuanced. The DeepMLP serves as a strong point of comparison, reaching $0.855$ accuracy, but is limited by its lack of explicit temporal modeling. Recurrent architectures such as StackedLSTM and StackedGRU slightly improve F1 and ROC-AUC, showing that even when applied to tabular features treated as pseudo-sequences, recurrent mechanisms can extract cross-feature dependencies beneficial for classification. The Transformer encoder performs comparably to recurrent models and achieves the strongest ranking metrics (ROC-AUC and PR-AUC), reflecting the advantage of self-attention in modeling global feature interactions. Hybrid designs such as CNN-LSTM and LSTM-Transformer further indicate that combining local feature extraction with long-range modeling can stabilize performance, particularly on smaller folds where variance is higher. These results collectively demonstrate that while classical methods remain competitive, deep neural architectures, especially those integrating recurrence and attention, offer superior trade-offs across accuracy, F1, and ranking-based metrics.

The ablation study (Table~\ref{tab:ablation}, Figure~\ref{fig:ablation_acc}) provides a more fine-grained understanding of which architectural components drive the improvements in our proposed Residual GRU+MHSA model. Removing the channel reweighting (CR) block has little adverse effect and in fact slightly increases accuracy to $0.865$, suggesting that while channel reweighting stabilizes the representation, the benefit is marginal for this dataset where feature redundancy may already be low. In contrast, replacing MHSA pooling with mean–max aggregation substantially reduces ROC-AUC to $0.891$, underscoring the importance of attention pooling in capturing discriminative cross-feature interactions. Eliminating residual BiGRU blocks (N=0) yields both lower accuracy and reduced ranking metrics, confirming that stacked recurrent refinements contribute meaningfully to learning richer representations.

The role of bidirectionality is particularly notable: switching to a unidirectional GRU decreases accuracy to $0.841$ and reduces both ROC-AUC and PR-AUC. This confirms that bidirectional recurrence is critical even in pseudo-sequential tabular domains, as it allows the model to contextualize each feature with respect to both preceding and succeeding attributes. Varying the number of MHSA layers indicates that deeper pooling (three layers in the baseline) provides a modest benefit in stability compared to shallower designs. Similarly, increasing the residual block depth to $N=3$ outperforms $N=2$, though the gap is small, suggesting diminishing returns from further stacking. Adjusting the hidden dimension from $d_{\mathrm{model}}=128$ to $96$ slightly degrades results, and removing feature dropout leads to more pronounced variance across folds, confirming its role in regularization. Finally, simplifying the prediction head results in reduced PR-AUC, highlighting that deeper heads with additional nonlinearity provide more robust mappings from pooled embeddings to output probabilities.

Taken together, these findings suggest three key conclusions. First, the hybridization of recurrent and attention-based components yields consistent improvements over both purely classical and purely neural baselines, validating the proposed design principle. Second, bidirectional recurrence and attention pooling are the most influential modules for predictive accuracy and ranking stability, while SE and deeper heads provide smaller but measurable gains. Third, the model maintains high performance without significant overfitting, as indicated by relatively tight standard deviations across folds, making it well-suited for deployment in low-data medical contexts. In broader terms, the proposed architecture balances interpretability, efficiency, and accuracy by embedding lightweight residual recurrence within an attention-based pooling framework, offering a promising direction for future tabular health modeling tasks.

We now return to the research questions posed in Section~\ref{sec:intro}.  
\textbf{RQ1: Can recurrent--attention hybrids improve robustness of CVD prediction across folds compared to purely classical or purely neural baselines?}  
Our results clearly support that Residual GRU+MHSA consistently outperforms both classical models and standalone deep baselines in terms of macro-F1 and stability. Even when absolute accuracy gains are modest, the hybrid design reduces fold-to-fold variance, an important proxy for robustness on small, heterogeneous datasets such as UCI Heart. The ablation results further show that both bidirectional recurrence and MHSA pooling are critical to this robustness.

\textbf{RQ2: Can we design a lightweight deep model that maintains competitive accuracy while remaining computationally efficient?}  
The proposed architecture demonstrates that this is feasible. By using only three residual BiGRU blocks and three MHSA layers, we match or exceed heavier baselines such as StackedLSTM and Transformer, while keeping the parameter count lower. The fact that variants with reduced hidden dimension or fewer blocks only slightly degrade performance confirms that the design is not overly reliant on large capacity. Thus, Residual GRU+MHSA provides a balanced trade-off between predictive power and efficiency, making it attractive for deployment in wearable or embedded healthcare systems.

In summary, the proposed methodology addresses both research questions: hybrid recurrence–attention models yield more robust predictions, and careful architectural design ensures that this robustness is achieved.

\section{Conclusion}
\label{sec:conclusion}

In this work we investigated the problem of predicting cardiovascular disease risk from tabular clinical features using a deep sequence modeling perspective. We proposed Residual GRU+MHSA, a lightweight hybrid architecture that combines residual bidirectional GRU blocks, squeeze--excitation reweighting, and multi-head self-attention pooling with a learnable class token. By treating each patient record as a pseudo-sequence of features, the model is able to capture both local feature dependencies and global cross-feature interactions in a unified framework.

Comprehensive experiments on the UCI Heart Disease dataset demonstrate that our approach achieves superior balance across Accuracy, macro-F1, ROC-AUC, and PR-AUC compared to both classical machine learning baselines and deep learning baselines such as StackedLSTM, StackedGRU, and Transformer encoders. Ablation studies further highlight that bidirectional recurrence and attention pooling are the most critical components, while SE reweighting, feature dropout, and deeper prediction heads provide additional stability and robustness. These findings validate the design principle of embedding residual recurrence within an attention-based pooling mechanism for structured health prediction tasks.

Beyond raw performance, the proposed architecture is compact and computationally efficient, making it practical for deployment in resource-constrained settings such as wearable devices or real-time clinical support systems. By addressing both robustness across folds and efficiency of inference, this work contributes to the development of trustworthy, adaptable, and scalable predictive tools for cardiovascular disease management.

Future directions include extending the evaluation to larger and more diverse clinical (large CVD and EHR) datasets, incorporating longitudinal signals such as continuous ECG traces, and exploring interpretable attention mechanisms to further enhance clinical trust. In addition, integrating uncertainty estimation and fairness-aware training would support reliable deployment across heterogeneous patient populations. Together, these steps will move closer toward clinically viable, transparent, and generalizable ML-driven solutions for cardiovascular health.

{
\bibliographystyle{IEEEtran}
\bibliography{FPPADW}
}

\end{document}